\documentclass[letterpaper]{article} 
\usepackage{aaai2027}  
\usepackage[hyphens]{url}  
\usepackage{graphicx} 
\usepackage{natbib}  
\usepackage{caption} 
\usepackage{algorithm}
\usepackage{algorithmic}
\usepackage{amsmath}
\usepackage{amssymb}
\usepackage{bm}
\usepackage{booktabs}

\usepackage{xcolor}

\usepackage{multirow}
\usepackage[table]{xcolor}
\usepackage{adjustbox}

\usepackage{xspace}
\newcommand{\system}{\texttt{MaP}\xspace}
\newcommand{\ie}{{\em i.e., \xspace}}

\newcommand{\eg}{{\em e.g., \xspace}}

\newcommand{\squishlist}{
\begin{list}{$\bullet$}{
  \setlength{\itemsep}{0pt}
  \setlength{\parsep}{3pt}
  \setlength{\topsep}{3pt}
  \setlength{\partopsep}{0pt}
  \setlength{\leftmargin}{3.5mm}
  \setlength{\labelwidth}{1em}
  \setlength{\labelsep}{0.5em}}}
\newcommand{\squishend}{\end{list}}

\title{Motion-as-Prompt: Enhancing Motion Reasoning in Multimodal Large Language Models via Motion-Guided Cross-Frame Visual Prompting
\thanks{Preprint. This manuscript is currently under review.}
}
\author{
	Xikai Sun\textsuperscript{\rm 1},
    Kebin Liu\textsuperscript{\rm 1},
    Haotian Wang\textsuperscript{\rm 2}, 
    Li Liu\textsuperscript{\rm 1}, 
    Xu Wang\textsuperscript{\rm 1}, 
    Yunhao Liu\textsuperscript{\rm 1} 
}
\affiliations{
   \textsuperscript{\rm 1}Tsinghua University, Beijing, China  \quad    
    \textsuperscript{\rm 2}JD Logistics, Beijing, China 
}

\begin{document}
	
	\maketitle
	
	\begin{abstract}
    Motion-centric video reasoning is fundamental to interactive applications such as robotic manipulation and autonomous navigation. However, multimodal large language models (MLLMs) typically process videos through sparse uniform sampling to control visual-token and attention costs. This strategy may discard critical transitions between sampled frames, limiting reasoning about object movement, collisions, and causal interactions. To mitigate this issue, we propose \textbf{Motion-as-Prompt (\system)}, a track-guided cross-frame visual prompting framework. \system recovers dense point trajectories, selects motion-informative frames, and marks the trajectories accumulated between consecutive sampled frames directly onto the visual inputs, making otherwise hidden displacement, direction changes, and interactions observable to frozen MLLMs. Experiments on CLEVRER and Something-Something-v2 show that \system consistently improves average motion-reasoning accuracy, yielding gains of $4.2\%$ and $8.9\%$ for GPT-5.5, respectively. 
    Notably, these improvements are obtained without degrading non-motion understanding, highlighting the robustness of \system.
    These results demonstrate that \system provides a simple and effective solution for enhancing motion-centric video reasoning without model training or architectural modification. Project page: \url{https://github.com/SunVictor23/MaP}.
    
	\end{abstract}

\section{Introduction}

Motion-centric video reasoning is a fundamental capability of multimodal large language models (MLLMs). 
It requires not only recognizing objects and scenes, but also understanding how entities move, interact, and cause state changes over time. 
This capability underpins a wide range of safety-critical and interactive applications, such as robotic manipulation, autonomous navigation, and augmented-reality assistance, where decisions depend on accurately perceiving trajectories, contacts, direction changes, and short-lived events. 
Although recent MLLMs have achieved strong performance on general video understanding~\cite{p_videochatgpt_maaz2024video,p_videochat_li2025videochat,p_videollama_cheng2024videollama,p_videollava_lin2024video}, their motion reasoning remains constrained by the way videos are presented to the model.

\begin{figure}
    \centering
    \includegraphics[width=1\linewidth]{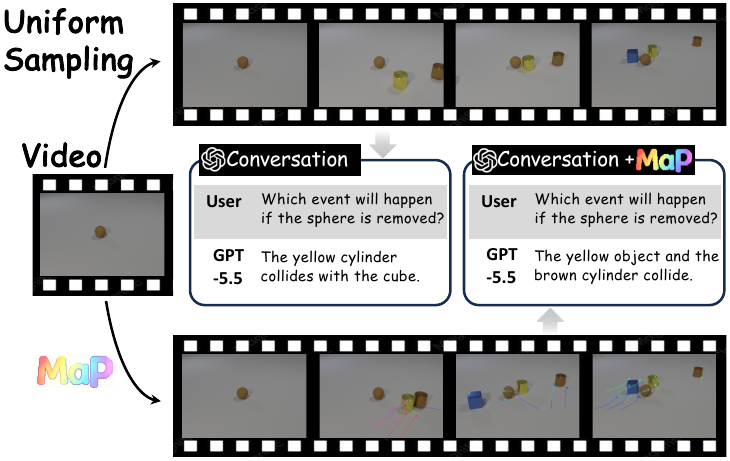}
    \caption{Illustration of inter-frame motion loss and its recovery with \system. Uniform sampling retains disconnected snapshots, while \system makes the missing motion observable.}
    \label{fig:figure1}
\end{figure}

To control visual-token and attention costs, MLLMs typically process only a sparse subset of video frames~\cite{p_longvu_shen2024longvu,p_aks_tang2025adaptive} by uniform sampling. While computationally efficient, sparse sampling converts a continuous motion process into a sequence of snapshots. 
Critical transitions occurring between sampled frames, such as acceleration, direction changes, and collisions, may therefore become completely invisible to the MLLM. 
This information bottleneck is referred to as inter-frame motion loss in our paper.
Figure~\ref{fig:figure1} illustrates this limitation. Sparse uniform sampling may preserve observations before and after an interaction while omitting the collision event between them. The MLLM must then infer the event from incomplete evidence and may consequently produce an incorrect prediction. 

Existing approaches do not directly mitigate this inter-frame motion loss. Motion-aware video models improve temporal perception by introducing specialized motion representations, architectural components, or task-specific fine-tuning~\cite{p_imove_li2025imove,p_vtimellm_huang2024vtimellm,p_videollama_cheng2024videollama}. 
However, these approaches require additional training or access to the model architecture. 
Visual prompting provides a more flexible alternative by augmenting the model's visual input with explicit guidance~\cite{p_exploring_zhang2024exploring}. Yet existing pixel-level prompts are predominantly frame-local: they annotate object regions, identities, or spatial relations within individual frames~\cite{p_som_yang2023set,p_gom_frisoni2026graph}. 
Such prompts help the model understand \emph{what} and \emph{where}, but do not reveal \emph{how} objects move across the unsampled intervals. 
This creates a fundamental mismatch, \ie the missing evidence is cross-frame, whereas existing visual prompts primarily describe sampled single-frame semantics.

Our key observation is that the discarded motion can be recovered from the original video and re-encoded into the sparse visual inputs. Based on this insight, we propose \textbf{Motion-as-Prompt (\system)}, a motion-guided cross-frame visual prompting framework for motion-centric video understanding. 
\system recovers dense point trajectories from the video, adaptively selects frames according to motion intensity, and explicitly marks the trajectories accumulated between consecutive sampled frames directly onto the visual inputs. 
In this way, \system transforms otherwise hidden temporal transitions into explicit visual prompts that can be interpreted by a frozen MLLM. The resulting framework is training-free, model-agnostic, and requires no architectural modification.

Our contributions are summarized as follows:
\begin{itemize}
    \item We identify inter-frame motion loss as an input-side bottleneck of sparsely sampled MLLMs and introduce cross-frame motion-recovery prompting, which recovers discarded motion from the original video and makes it directly observable in sparse visual inputs.
    \item We instantiate this idea as Motion-as-Prompt (\system), a training-free framework that combines motion-guided sampling with inter-frame trajectory marking for motion-centric video reasoning.
    \item Experiments on motion benchmarks show that \system consistently improves the performance of MLLMs, with gains of $4.2\% \sim 8.9\%$ on GPT-5.5. Ablations further show that trajectory marking is the primary source of improvement and becomes more effective with larger frame budgets.
\end{itemize}
\section{Related Work}

\paragraph{Keyframe sampling.} MLLMs typically process only a small subset of frames, since dense video encoding incurs prohibitive visual-token and attention costs. Recent works aim to preserve the most informative frames under a fixed frame budget. 
For example, Adaptive Keyframe Sampling (AKS) balances query relevance and temporal coverage~\cite{p_aks_tang2025adaptive}. FOCUS and other learned selectors estimate frame importance from semantic cues~\cite{p_focus_zhu2025focus}. VideoTree organizes video evidence into a query-adaptive hierarchy~\cite{p_videotree_wang2025videotree}. MDP3 jointly accounts for relevance, diversity, and temporal order~\cite{p_mdp3_sun2025mdp3}.
Token-compression methods such as LongVU~\cite{p_longvu_shen2024longvu}, Dynamic-VLM~\cite{p_dynamicvlm_wang2025dynamic}, and FastVID~\cite{p_fastvid_shen2026fastvid} further reduce redundant frames or spatial tokens to support longer inputs. 
Despite their differences, they largely treat sparse sampling as an evidence-selection problem, and their decisions rest on semantic relevance, visual similarity, or token redundancy. Consequently, they may retain frames that describe what appears in a video, but discard how objects move between frames.

\paragraph{Spatiotemporal modeling and visual prompting.} Another line of work enhances the spatiotemporal reasoning of MLLMs through model architecture~\cite{p_videoglamm_munasinghe2025videoglamm}, supervision, or explicit motion features~\cite{p_efficient_zhao2025efficient, p_remora_yashima2026remora}. TimeChat binds visual content to timestamps~\cite{p_timechat_ren2024timechat}, VTimeLLM introduces temporal-boundary-aware training~\cite{p_vtimellm_huang2024vtimellm}, and Seq2Time transfers sequential knowledge to temporal grounding~\cite{p_seq2time_deng2025seq2time}. SlowFocus combines low-frequency global observation with high-frequency sampling of query-relevant clips~\cite{p_slowfocus_nie2024slowfocus}, while Time-R1 improves temporal grounding through task-specific post-training~\cite{p_time-r1_wang2026time}. 
Although these methods acknowledge that the motion evidence supplied by keyframes alone is insufficient, they typically require modifying models, using specialized video representations, or fine-tuning. 
Visual prompting offers a lightweight alternative. It can operate in the embedding space through learnable prompt tokens~\cite{p_visual_jia2022visual}, or directly in the pixel space through visible points, boxes, masks, and scribbles~\cite{p_vip_cai2024vip}.
Set-of-Mark (SoM) supports visual grounding by overlaying segmented regions and numeric identifiers~\cite{p_som_yang2023set}, while Graph-of-Mark (GoM) encodes objects and their spatial relations as a visible scene graph~\cite{p_gom_frisoni2026graph}. ViKey uses visible frame identifiers as temporal anchors~\cite{p_vikey_lee2026vikey}, and STOP introduces spatial and temporal prompts to emphasize discriminative regions~\cite{p_stop_liu2025stop}. 
Those visual-prompting methods inject object indices, temporal panels, or learned spatiotemporal prompts to improve temporal correspondence. However, they mainly encode intra-frame semantics in individual frames, \eg object identity, region membership, or spatial layout. \system complements this line: it selects motion-informative keyframes and recovers inter-frame motion information, making it observable and usable by the MLLM.

\paragraph{Motion-aware video understanding.}
Recent studies explicitly incorporate motion cues into MLLMs rather than relying solely on frame-level semantics. 
VideoExpert processes high-frame-rate compressed features with a dedicated temporal expert to capture dynamic variations~\cite{p_videoexpert_zhao2026videoexpert}. 
Flow4Agent introduces optical-flow priors for temporal content organization and motion-aware token pruning~\cite{p_flow4agent_liu2025flow4agent}. 
DynImg uses non-key frames as temporal prompts to highlight regions containing rapid motion~\cite{p_dynimg_bao2025dynimg}, while MotionSight employs object-centric spotlight and motion-blur prompts for zero-shot fine-grained motion understanding~\cite{p_motionsight_du2025motionsight}. 
Unlike these methods, \system reconstructs explicit point trajectories from the full-frame-rate video and directly renders the motion accumulated between consecutive sampled frames, without modifying or training the MLLM. 
This ensures the generalization ability of \system across diverse scenarios.
\section{Method}

\begin{figure*}[t]
    \centering
    \includegraphics[width=1.8\columnwidth]{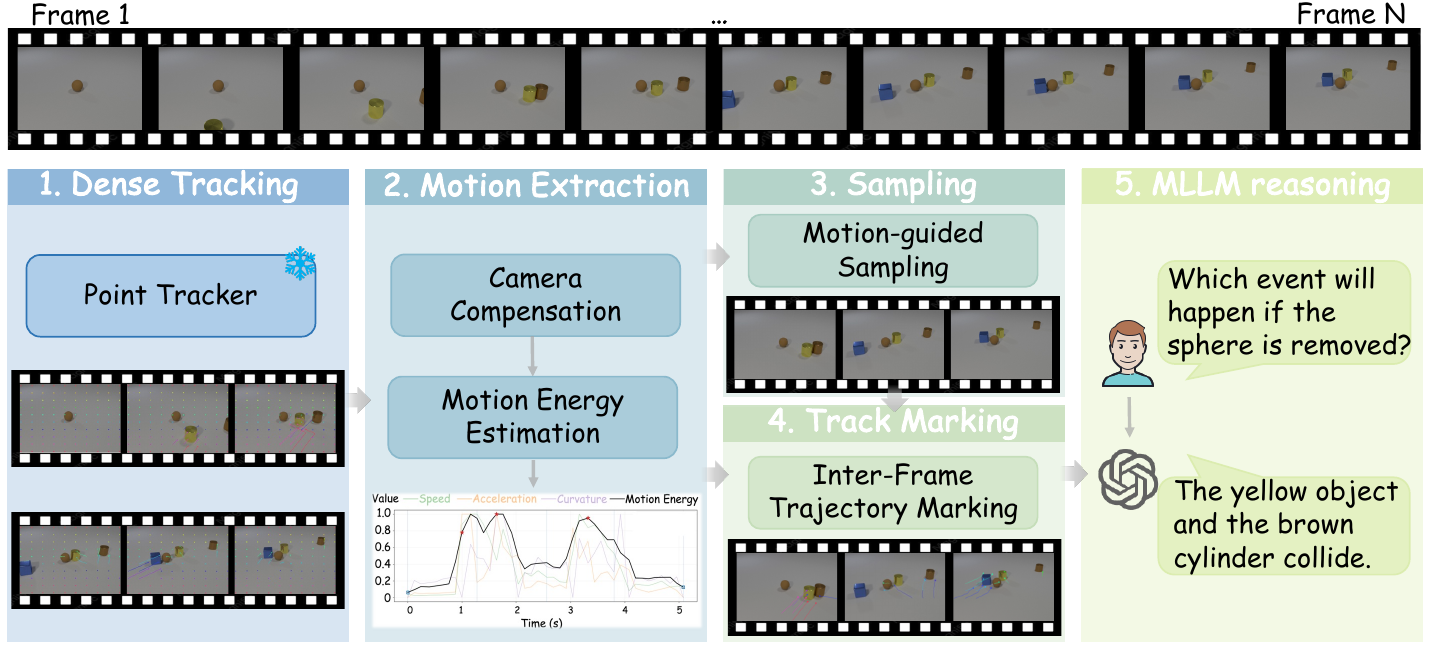}
    \caption{Overview of \system. Given a full-frame-rate video, \system first recovers dense point trajectories using a frozen point tracker. It then compensates for global camera motion, estimates frame-wise motion energy, and selects motion-informative frames under a fixed frame budget. Finally, trajectory segments accumulated between consecutive sampled frames are rendered onto the later frames and provided to the frozen MLLM for motion-aware reasoning.}
    \label{fig:pipeline}
\end{figure*}

In this section, we present \system, which aims to shift an MLLM's perception from reasoning across disconnected snapshots to a direct understanding of motion. A compact algorithmic overview of \system, expressed as pseudocode, is provided in the Supplementary Material.

\subsection{Problem Definition and Overview of \system}

Given a video $V$ with $N$ frames at full frame rate and with a duration of $t$ seconds, and a text prompt $T$ containing the task instruction, an MLLM completes text as a response according to the distribution $P_{\text{MLLM}}(\cdot\mid F(V), T)$, where $F$ is the configured sampling rate. Recent studies show that strategically altering the visual tokens can indirectly steer the model's output. \cite{p_controlmllm_wu2024controlmllm}. Building on this premise, we propose \system as an operator acting on $V$ that produces an augmented frame sequence:
\begin{equation*}
    P_{\text{MLLM}}\big(\cdot\mid \text{\system}(V),\ T\big).
\end{equation*}
Without changing the total frame budget $B = F\cdot t$, $\text{\system}(V)$ produces two things: (i) a motion-aware set of selected frames $\mathcal{S}'$ ($|\mathcal{S}'| = B$), and (ii) a set of inter-frame motion marks $\mathcal{M}$ overlaid on those frames. 

The overview of \system is shown in Figure~\ref{fig:pipeline}. 
Given a motion-reasoning task, \system first employs a frozen point tracker to recover dense trajectories from the full-frame-rate video and aggregates them into motion-energy scores. These scores guide the adaptive selection of motion-informative frames $\mathcal{S}'$. \system then renders the inter-frame trajectories as visual markers $\mathcal{M}$ on the selected frames and feeds the augmented sequence into the MLLM for motion reasoning.

\subsection{Motion Signal Extraction}

\paragraph{Dense tracking.} To characterize the motion at each frame, we track a $G\times G$ grid of query points with a frozen point tracker $\mathcal{T}$ over the full-rate frames, re-seeding the grid at a fixed interval (\eg every second) to admit new objects that enter mid-clip~\cite{p_tapir_doersch2023tapir}. It yields point tracks $\mathbf{p}_\ell^i \in \mathbb{R}^2$ and visibility flags $o_\ell^i \in \{0,1\}$ for $\ell = 1,\dots,N$ frames and $i = 1,\dots,G^2$ query points.

\paragraph{Camera-compensated query point velocity.} The raw inter-frame displacement $\mathbf{p}_{\ell+1}^i-\mathbf{p}_\ell^i$ contains both the query point's object motion and camera self-motion. 
We model the latter using a global similarity transformation that best explains the displacement of all visible query points, including translation $\mathbf{b}$, rotation $\mathbf{R}$, and scale $s$, \ie $\mathbf{p}\mapsto s\mathbf{R}\mathbf{p}+\mathbf{b}$~\cite{p_action_wang2013action}. 
For efficient estimation, we parameterize the transformation by $\boldsymbol{\theta}=(a,b,t_x,t_y)$ (where $a=s\cos\phi,\, b=s\sin\phi$) and fit it over the set of points visible in both adjacent frames, \ie $\mathcal{V}_\ell=\{i: o_\ell^i o_{\ell+1}^i=1\}$:
\begin{equation*}
    \boldsymbol{\theta}_\ell^\star=\arg\min_{\boldsymbol{\theta}}\sum_{i\in\mathcal{V}_\ell}\big\|\mathbf{S}_{\boldsymbol{\theta}}(\mathbf{p}_\ell^i)-\mathbf{p}_{\ell+1}^i\big\|^2,
\end{equation*}
\begin{equation*}
    \mathbf{S}_{\boldsymbol{\theta}}(x,y)=\begin{bmatrix}ax-by+t_x\\ bx+ay+t_y\end{bmatrix},
\end{equation*}
which is a linear least-squares problem in $\boldsymbol{\theta}$ (closed-form, solved when $|\mathcal{V}_\ell|\ge 3$, else the camera term is $0$). The object velocity is the residual after removing the fitted camera flow:
\begin{equation*}
    \mathbf{v}_\ell^i=\big(\mathbf{p}_{\ell+1}^i-\mathbf{p}_\ell^i\big)-\big(\mathbf{S}_{\boldsymbol{\theta}_\ell^\star}(\mathbf{p}_\ell^i)-\mathbf{p}_\ell^i\big),\quad i\in\mathcal{V}_\ell.
\end{equation*}
For a static camera, $\mathbf{S}_{\boldsymbol{\theta}_\ell^\star}$ is approximately the identity transformation, and the compensation has little effect. For a moving camera, it suppresses camera-induced displacement, yielding a cleaner estimate of query points' motion.

\paragraph{Motion energy.} Given $\{\mathbf{v}_\ell^i\}$, we define a scalar motion energy score $M(\ell)$ for each frame. Specifically, we characterize the motion at frame $\ell$ along three complementary dimensions, including speed, acceleration, and curvature, as follows: 
\begin{equation*}
    \text{spd}_\ell=\frac{1}{n_\ell}\sum_{i=1}^{G^2} o_\ell^i\frac{\|\mathbf{v}_\ell^i\|}{d},\quad
    \text{acc}_\ell=\frac{1}{n_\ell}\sum_{i=1}^{G^2} o_\ell^i\frac{\|\mathbf{v}_\ell^i-\mathbf{v}_{\ell-1}^i\|}{d},
\end{equation*}
\begin{equation*}
    \text{cur}_\ell=\frac{1}{n_\ell}\sum_{i}^{G^2} o_\ell^i\,
    \arccos\!\frac{\langle \mathbf{v}_{\ell-1}^i,\mathbf{v}_\ell^i\rangle}{\|\mathbf{v}_{\ell-1}^i\|\,\|\mathbf{v}_\ell^i\|}\cdot
    \frac{\min(\|\mathbf{v}_{\ell-1}^i\|,\|\mathbf{v}_\ell^i\|)}{d},
\end{equation*}
where $n_\ell=\max(\sum_i o_\ell^i,1)$ is the number of visible points on the frame and $d=\sqrt{H^2+W^2}$ is the frame diagonal used for normalization. The turn angle in $\text{cur}_\ell$ is weighted by the smaller of the two adjacent velocities, preventing nearly static points from producing spurious curvature. Each descriptor is then normalized by its 95th percentile to reduce sensitivity to outliers, \ie $\widehat{u}=\operatorname{clip}(u/\operatorname{pct}_{95}(u),0,1)$ for $u\in\{\text{spd}_\ell,\text{acc}_\ell,\text{cur}_\ell\}$. 
The normalized descriptors are equally weighted and summed, smoothed using a length-$w$ moving-average operator $\Phi_w$ with $w=3$, and finally re-normalized as follows:
\begin{equation*}
    M(\ell)=\widehat{\Phi_w\!\big(\widehat{\text{spd}}_\ell+\widehat{\text{acc}}_\ell+\widehat{\text{cur}}_\ell\big)}\ \in[0,1].
\end{equation*}
$M(\ell)$ quantifies the intensity of motion at frame $\ell$, assigning higher scores to moments of rapid movement, acceleration, or directional change. These frames contain motion evidence that sparse sampling should preserve.

\subsection{Motion-Guided Sampling}

Given the motion-energy sequence $M(\cdot)$ over an $N$-frame video and a frame budget $B<N$, we introduce a motion-guided sampling mechanism to select an index set $\mathcal{S}'\subset\{1,\dots,N\}$, $|\mathcal{S}'|=B$. It consists of the following 3 stages.

\paragraph{Anchors for coverage.} 
To prevent selected frames from clustering around local motion peaks and ignoring the overall video,
we uniformly sample $n_a=\max(2,\lceil\alpha B\rceil)$ frames as anchors, where $0<\alpha<1$. Their indices are defined as:
\begin{equation*}
    \mathcal{A}=\big\{\operatorname{round}\!\big(\tfrac{k(N-1)}{n_a-1}\big):k=0,\dots,n_a-1\big\}.
\end{equation*}
These anchors preserve global temporal coverage.

\paragraph{Motion peaks with non-maximum suppression.} 
The remaining $B-|\mathcal{A}|$ frames are allocated to high-energy motion peaks using greedy non-maximum suppression (NMS). We impose a minimum temporal gap $r=\max(1,\lfloor N/(2B)\rfloor)$ between each candidate peak and all previously selected frames, including the anchors.
Specifically, candidate indices are examined in descending order of $M(\ell)$. An index $j$ is selected only if no selected index lies within $[j-r,j+r]$, after which this interval is suppressed. 
Because $r$ is approximately half the uniform sampling interval $N/B$,  peak frames may pack up to twice as densely as uniform samples, while avoiding excessive concentration at a single instant.
    
\paragraph{Importance-sampling fallback.} If NMS returns fewer than $B$ frames, as may occur when $M(\ell)$ is nearly flat, or the suppression intervals cover most candidates, the remaining slots are filled in descending order of motion energy without enforcing the gap constraint.

Finally, the selected indices are mapped to timestamps for constructing the visual prompt.
The complete motion-guided sampling is summarized as pseudocode in Algorithm~\ref{alg:sampling}.

\begin{algorithm}[tb]
    \caption{Motion-Guided Sampling}
    \label{alg:sampling}
    \begin{algorithmic}[1]
        \REQUIRE Motion energy scores $M\in[0,1]^N$; frame budget $B$; anchor ratio $\alpha$\\
        \ENSURE selected frame indices $\mathcal{S'}$
        \STATE $r \leftarrow \max(1,\lfloor N/(2B)\rfloor)$
        \STATE $n_a \leftarrow \max(2,\lceil\alpha B\rceil)$
        \STATE $\mathcal{S}' \leftarrow \textsc{Uniform}(n_a)$ \\
        \COMMENT{Motion-peak selection with temporal NMS}
        \STATE block interval $[j-r,\,j+r]$ for all $j\in\mathcal{A}$ 
        \FOR{$j$ in $\operatorname{argsort}\!\downarrow\! M$ \textbf{while} $|\mathcal{S'}|<B$}
        \IF{$j$ is unblocked}
        \STATE $\mathcal{S'} \leftarrow \mathcal{S'}\cup\{j\}$;\quad block $[j-r,\,j+r]$
        \ENDIF
        \ENDFOR \\
        \COMMENT{Motion-energy fallback}
        \IF{$|\mathcal{S'}|<B$}
        \STATE add top-$M$ indices until $|\mathcal{S'}|=B$
        \ENDIF
        \STATE \textbf{return} $\operatorname{sort}(\mathcal{S'})$
    \end{algorithmic}
\end{algorithm}

%

\subsection{Inter-Frame Trajectory Marking}

After sampling the video frames, we introduce an inter-frame trajectory-marking mechanism that renders the query point motion between adjacent sampled frames onto the later frame, as a Set-of-Marks-style visual symbol.

\paragraph{Moving-point selection.}  To reduce visual clutter, we retain only query points with sufficiently large motion. For each grid point $i$ initialized on sampled frame $\ell$, we measure its maximum displacement over the following $\Delta$ frames. Point $i$ is retained if its maximum displacement exceeds a fraction $\tau=0.03$ of the frame width $W$, as follows:
\begin{equation*}
    \max_{m=1,\dots,\Delta}\big\|\mathbf{p}_{\ell+m}^i-\mathbf{p}_\ell^i\big\| \;>\; \tau\,W.
\end{equation*}
The retained points are ordered by total motion, and at most $K$ points are kept to control annotation density.

\paragraph{Inter-frame trajectory rendering.} Let the sampled frame indices be $\mathcal{S}'=\{s_1<\dots<s_B\}$. On the $s_k$-th  frame, for each retained visible point $i$, we render only the trajectory segment $\gamma_{k}^i=\big(\mathbf{p}_\ell^i\big)_{\ell=s_{k-1}}^{s_k}$ accumulated since the  $s_{k-1}$-th frame.
Each trajectory is drawn as a single-colored polyline, with a circular marker at its endpoint $\mathbf{p}_{s_k}^i$. 
By constructing the inter-frame trajectories as marks, \system explicitly fuses the missing inter-frame motion into the sampled frames.
Together with a corresponding text prompt, \system guides the MLLM to exploit the motion cues during motion reasoning.

\section{Experiments}

\begin{table*}[t]
    \centering
    \setlength{\tabcolsep}{9pt}
    
    \begin{tabular}{clccccccccc}
        \toprule
        \multirow{2}{*}{\textbf{MLLM}}
        & \multirow{2}{*}{\textbf{Method}}
        & \multirow{2}{*}{\textbf{Sel.}}
        & \multirow{2}{*}{\textbf{Mark.}}
        & \multicolumn{6}{c}{\textbf{CLEVRER}}
        & \multirow{2}{*}{\textbf{SSv2}} \\
        \cmidrule(lr){5-10}
        & & & & \textbf{OE} & \textbf{MD} & \textbf{MC} & \textbf{MA} & \textbf{CI} & \textbf{Avg.} & \\
        \midrule

        \multirow{6}{*}{
            \rotatebox[origin=c]{90}{\textbf{Qwen3-VL-2B}}
        }
        & uniform
        & $\times$ & $\times$
        & 63.1 & \textbf{40.5} & 53.5 & 76.0 & 55.0 & 57.6 & 51.8 \\

        & AKS
        & sem. & $\times$
        & 62.6 & 27.5 & 50.0 & 62.5 & 50.0 & 50.5 & 53.3 \\

        & FOCUS
        & sem. & $\times$
        & 57.1 & 31.5 & 47.5 & 66.0 & 56.5 & 51.7 & \textbf{57.2} \\

        & SoM
        & $\times$ & sem.
        & 51.0 & 26.5 & 34.5 & 55.0 & 38.0 & 41.0 & 41.5 \\

        & GoM
        & $\times$ & sem.
        & 58.6 & 34.5 & 51.0 & 70.5 & 44.5 & 51.8 & 42.0 \\

        & \textbf{\system (ours)}
        & \textbf{mot.} & \textbf{mot.}
        & \textbf{69.2} & \textbf{40.5} & \textbf{54.0} & \textbf{79.5} & \textbf{59.0} & \textbf{60.4} & 53.2 \\

        \midrule

        \multirow{6}{*}{
            \rotatebox[origin=c]{90}{\textbf{GPT-5.5}}
        }
        & uniform
        & $\times$ & $\times$
        & \textbf{87.9} & 72.5 & 69.5 & 81.5 & 63.0 & 74.9 & 71.1 \\

        & AKS
        & sem. & $\times$
        & 73.2 & 57.5 & 49.5 & 69.0 & 57.5 & 61.4 & 67.9 \\

        & FOCUS
        & sem. & $\times$
        & 77.8 & 67.5 & 61.0 & 74.0 & 59.0 & 67.9 & 77.6 \\

        & SoM
        & $\times$ & sem.
        & 68.7 & \textbf{84.5} & 44.0 & 63.5 & 48.5 & 61.8 & 66.1 \\

        & GoM
        & $\times$ & sem.
        & 78.8 & 70.5 & 66.0 & 76.5 & 62.0 & 70.8 & 67.1 \\

        & \textbf{\system (ours)}
        & \textbf{mot.} & \textbf{mot.}
        & 87.4 & 75.5 & \textbf{72.0} & \textbf{85.0} & \textbf{75.5} & \textbf{79.1}  & \textbf{80.0} \\

        \bottomrule
    \end{tabular}
    \caption{Accuracy comparison with baselines on CLEVRER and SSv2.
    ``Sel.'' and ``Mark.'' denote frame selection and visual prompt marking. ``sem.'' and ``mot.'' denote semantic and motion cues, respectively.}
    \label{tab:main}
\end{table*}

\subsection{Experimental Setup}

\paragraph{Benchmarks and metrics.} 
We evaluate \system on two complementary motion-reasoning benchmarks: CLEVRER~\cite{p_clevrer_yi2019clevrer}, which focuses on motion-intensive reasoning in synthetic scenes, and Something-Something-v2 (SSv2)~\cite{p_ssv2_goyal2017something}, which evaluates fine-grained human–object interactions in real-world videos.
CLEVRER comprises five sub-tasks (as defined in MVBench): object existence (OE), moving direction (MD), moving count (MC), moving attribute (MA), and counterfactual inference (CI)~\cite{p_mvbench_li2024mvbench}.  
For SSv2, we evaluate on its validation set using a four-way multiple-choice formulation in which all object references in the answer options are replaced with ``something.'' 
This abstraction requires the model to identify the underlying action rather than rely on object recognition.
In addition, we test on the video-reasoning benchmark TempCompass~\cite{p_tempcompass_liu2024tempcompass}, whose tasks are not about motion, to examine the generality of our method. Accuracy is reported for all tasks.

\paragraph{Baselines.} We compare against two representative families of training-free methods. For keyframe selection, AKS and FOCUS select query-relevant frames based on semantic cues. For pixel-level visual prompting, SoM overlays object segmentation masks and indices on the video frames, whereas GoM renders objects and their spatial relations as an in-frame scene graph.

\paragraph{Implementation.} We evaluate \system and the baselines on Qwen3-VL-2B-Instruct (local weights)~\cite{p_qwen3vl_bai2025qwen3}, and GPT-5.5 (remote, OpenAI-compatible API)~\cite{p_gpt5-5_openai2026gpt55}. Both models are frozen. The point tracker is CoTracker3, tracking a $10\times10$ grid of query points. Anchor sampling defaults to $\alpha=1/4$ with NMS radius $r=\max(1, \lfloor N/2B\rfloor)$. The sampling rate is uniformly set to $1\ \mathrm{FPS}$.


In our experiments, we aim to answer:
\textbf{Q1 (\S4.2):} How does \system perform on motion-reasoning tasks?
\textbf{Q2 (\S4.3):} Does the motion-guided sampling harm non-motion video-reasoning tasks?
\textbf{Q3 (\S4.4):} How sensitive is the gain of inter-frame trajectory marking to the frame budget?
\textbf{Q4 (\S4.5):} What is the marginal contribution of each component?
\textbf{Q5 (\S4.6):} Is \system's preprocessing overhead acceptable?


\subsection{Results on Motion Benchmarks}
\label{sec:benchmark_results}

\begin{figure*}[tb]
    \centering
    \includegraphics[width=1\linewidth]{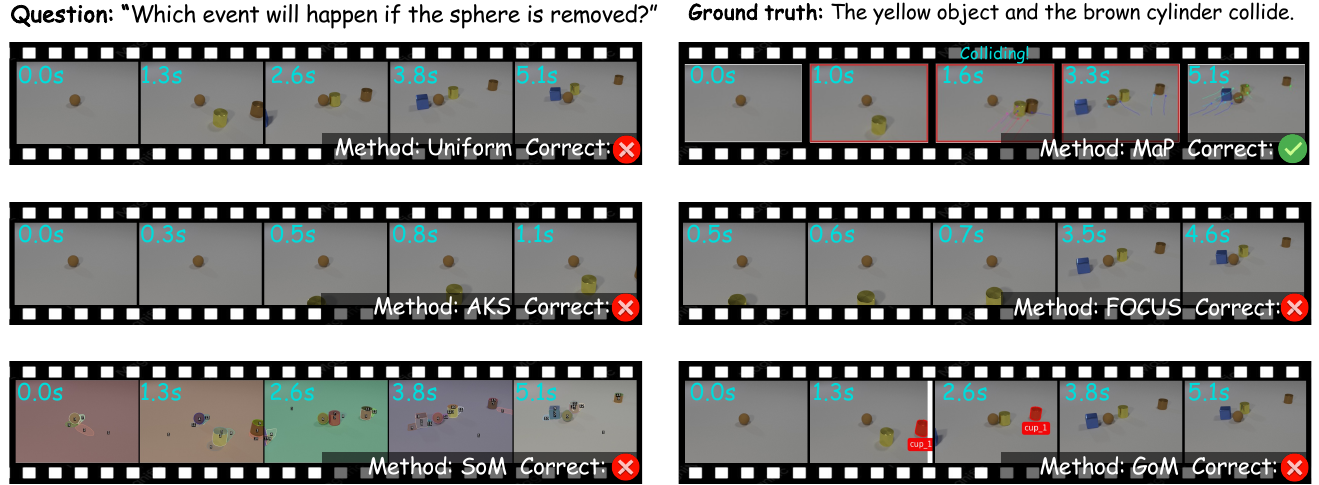}
    \caption{Qualitative comparison on a 5.1-second CLEVRER video sampled at $1\ \mathrm{FPS}$ (\ie five-frame budget). Red boxes highlight the keyframes selected by \system's motion-guided sampler.}
    \label{fig:compare}
\end{figure*}

We compare \system against the base model and four representative baselines on two motion-reasoning benchmarks, as shown in Table~\ref{tab:main}. 
On CLEVRER, \system achieves the highest average accuracy for both MLLMs, indicating that the models can effectively exploit trajectory marks to recover object motion from sparsely sampled frames. The larger improvement on GPT-5.5 further suggests that stronger models are better able to integrate the recovered motion cues into their reasoning.
In contrast, semantic keyframe selectors such as AKS and FOCUS substantially degrade performance on CLEVRER. The critical evidence in this benchmark often lies in continuous transitions whose frame-level semantics vary only slightly, rather than in a few semantically salient snapshots. Consequently, non-uniform sampling may omit short-lived events such as collisions and direction changes. It also produces irregular temporal intervals between frames, making motion more difficult to infer. Similarly, the dense static annotations introduced by SoM and GoM obscure task-relevant visual evidence and generally reduce performance.
Figure~\ref{fig:compare} compares the visual inputs produced by \system and the baselines. In this example, correct reasoning requires preserving both the critical collision moment and its surrounding motion. AKS and FOCUS miss the event because the semantic changes between adjacent frames are subtle, while the annotations introduced by SoM and GoM interfere with the MLLM's reasoning. In contrast, \system preserves the collision and explicitly visualizes the associated inter-frame motion, leading to the MLLM's correct prediction.

An exception appears in the moving-direction task on GPT-5.5, where SoM achieves the highest score (84.5\% versus 75.5\% for \system). This task asks about the movement of a single specified object (\eg ``which way is the cyan sphere moving''), and SoM's segmentation-and-index annotations make that object easy to localize consistently across frames. However, such object-centric identity cues provide less benefit for other tasks, resulting in a lower overall average. This observation also motivates the object-abstracted SSv2 formulation, which reduces the possibility of solving the task through object recognition or explicit identity cues rather than motion understanding.

To assess generalization beyond synthetic scenes, we further evaluate \system on SSv2. On GPT-5.5, \system achieves the best accuracy of 80.0\%, outperforming uniform sampling by 8.9\% and the strongest baseline, FOCUS, by 2.4\%. This result is consistent with CLEVRER, showing that a capable model can effectively exploit the recovered motion cues in real-world videos.
On Qwen3-VL-2B, \system improves over the base model (53.2\% versus 51.8\%), but remains below FOCUS (57.2\%) and comparable to AKS (53.3\%). We attribute this to the nature of SSv2, whose choices often depend on recognizing action semantics, such as ``pushing something from right to left,'' which directly favors the query-relevance signals used by AKS and FOCUS. In contrast, interpreting trajectory overlays in visually dense scenes places greater demands on the model's perceptual and reasoning capacity.

\subsection{Generalization to Natural Videos}



TempCompass does not specifically target object-motion reasoning, and we use it to evaluate whether motion-guided sampling transfers robustly to broader video-reasoning tasks, as shown in Table~\ref{tab:tempcompass}. Due to anchors for coverage, \system's sampling mechanism matches or slightly improves performance on both MLLMs, achieving average accuracy of 67.6\% versus 67.3\% on Qwen3-VL-2B and 88.6\% versus 88.3\% on GPT-5.5. It also yields small gains on most individual tasks, whereas the other baselines generally reduce performance for the reasons discussed in \ref{sec:benchmark_results}. These results suggest that motion-guided sampling does not compromise performance on non-motion-oriented video-reasoning tasks.

\begin{table}[t]
    \centering
    \small
    \setlength{\tabcolsep}{7pt}
    \begin{tabular}{@{}clccccc@{}}
        \toprule
        MLLM & Method & MC & Y/N & CM & Cap. & Avg. \\
        \midrule
        \multirow{6}{*}{\rotatebox[origin=c]{90}{\textbf{Qwen3-VL-2B}}}
        & uniform & 65.8 & \textbf{68.8} & 76.8 & 57.8 & 67.3 \\
        & AKS & 52.6 & 61.4 & 66.9 & 46.8 & 56.9 \\
        & FOCUS & 61.1 & 65.1 & 73.5 & 49.6 & 62.3 \\
        & SoM & 55.6 & 61.1 & 68.6 & 43.3 & 57.2 \\
        & GoM & 60.6 & 63.0 & 70.5 & 47.6 & 60.4 \\
        & \textbf{Mot. (ours)} & \textbf{66.4} & \textbf{68.8} & \textbf{77.2} & \textbf{57.9} & \textbf{67.6} \\
        \midrule
        \multirow{6}{*}{\rotatebox[origin=c]{90}{\textbf{GPT-5.5}}}
        & uniform & 85.9 & 87.4 & \textbf{92.3} & \textbf{87.8} & 88.4 \\
        & AKS & 68.5 & 73.1 & 79.6 & 63.2 & 71.1 \\
        & FOCUS & 81.5 & 82.1 & 87.8 & 78.1 & 82.4 \\
        & SoM & 77.4 & 79.8 & 87.1 & 78.6 & 80.7 \\
        & GoM & 81.8 & 82.5 & 88.5 & 83.3 & 84.0 \\
        & \textbf{Mot. (ours)} & \textbf{86.2} & \textbf{88.2} & \textbf{92.3} & 87.7 & \textbf{88.6} \\
        \bottomrule
    \end{tabular}
    \caption{Accuracy comparison on TempCompass, which consists of four tasks. MC, Y/N, CM, and Cap. denote multiple-choice, yes/no, caption matching, and captioning, respectively. ``Mot.'' denotes motion-guided sampling of \system.
    }
    \label{tab:tempcompass}
\end{table}

\subsection{Frame-Budget Analysis}

\begin{figure}[t]
    \centering
    \includegraphics[width=1\linewidth]{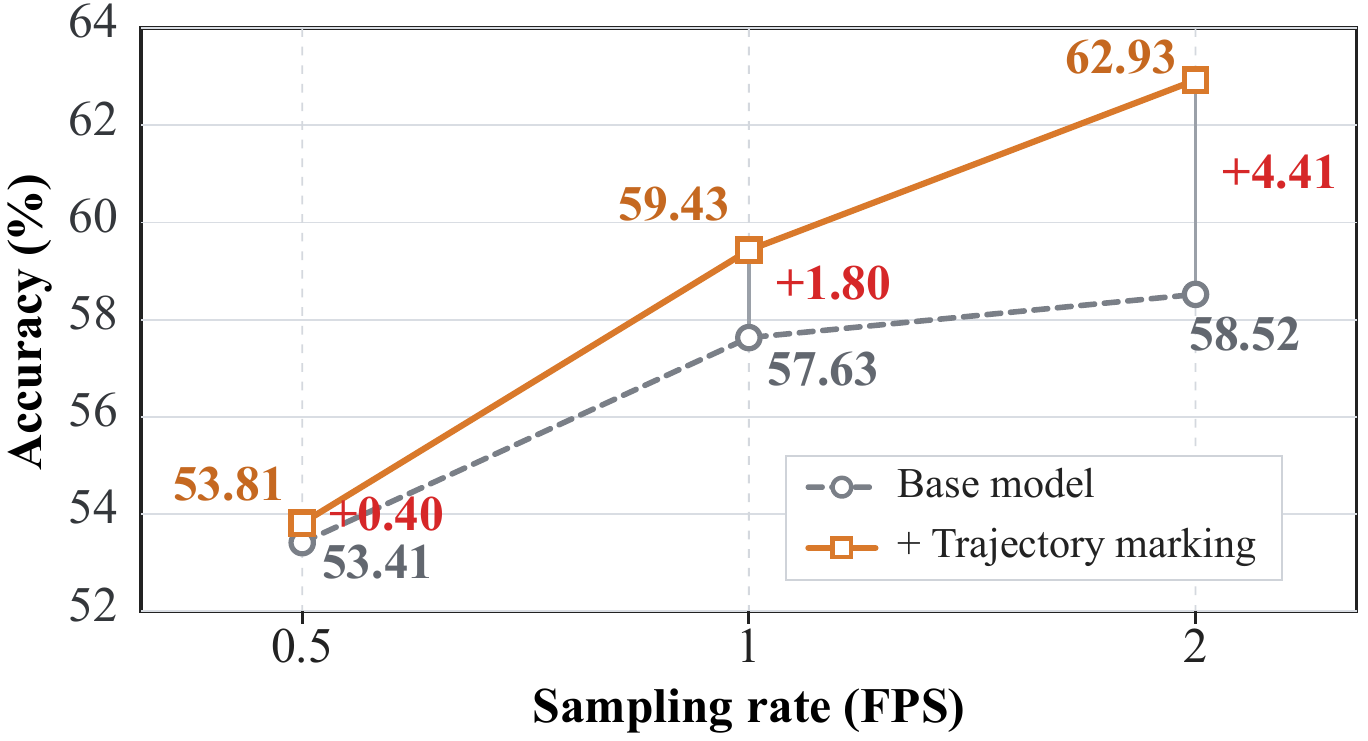}
    \caption{Trajectory-marking gains of Qwen3-VL-2B on CLEVRER under different frame budgets.}
    \label{fig:FPS_budget}
\end{figure}

Beyond motion-guided sampling, we examine how the effectiveness of trajectory marking varies with the frame budget. Specifically, we evaluate CLEVRER at three sampling rates, $\{0.5,1,2\} \mathrm{FPS}$. To isolate the contribution of marking, we fix the sampler to uniform sampling and vary only whether trajectory marks are rendered, ensuring that both variants use identical frame positions. Figure~\ref{fig:FPS_budget} reports the marking gain in different FPS settings. The gain increases consistently with the sampling rate, indicating that trajectory marking becomes more effective when denser observations provide richer and more continuous motion evidence.


This result is initially counterintuitive. If sparse sampling discards more motion, one might expect trajectory marking to provide larger gains under smaller frame budgets. The opposite trend reveals that marking effectiveness depends not only on how much motion is missing, but also on how faithfully that motion can be represented between sampled frames. 
At $0.5\ \mathrm{FPS}$, some videos contain only two sampled frames, causing the inter-frame trajectory to collapse into a coarse stroke. Moreover, overlaying a mark on one of only a few frames can obscure raw visual evidence and compete for the model's attention. As the frame budget increases, the temporal interval between sampled frames becomes shorter, allowing each trajectory to capture motion with greater fidelity and better preserve curvature, direction changes, and variations in speed. The cost of visual overlay is also distributed across more frames.

Thus, trajectory marking does not simply restore a fixed amount of motion lost by sparse sampling. Its value depends on the fidelity of motion representations constructed between sampled frames, which improves with denser temporal observations. This finding suggests that a moderate frame budget with trajectory marking may offer a better trade-off than aggressive sparsity, improving motion reasoning while keeping the visual context manageable. Overall, the benefit of trajectory marking grows with the frame budget because it is governed by trajectory fidelity rather than frame scarcity alone.

\subsection{Ablation: Cumulative Gains}

\begin{figure}
    \centering
    \includegraphics[width=1\linewidth]{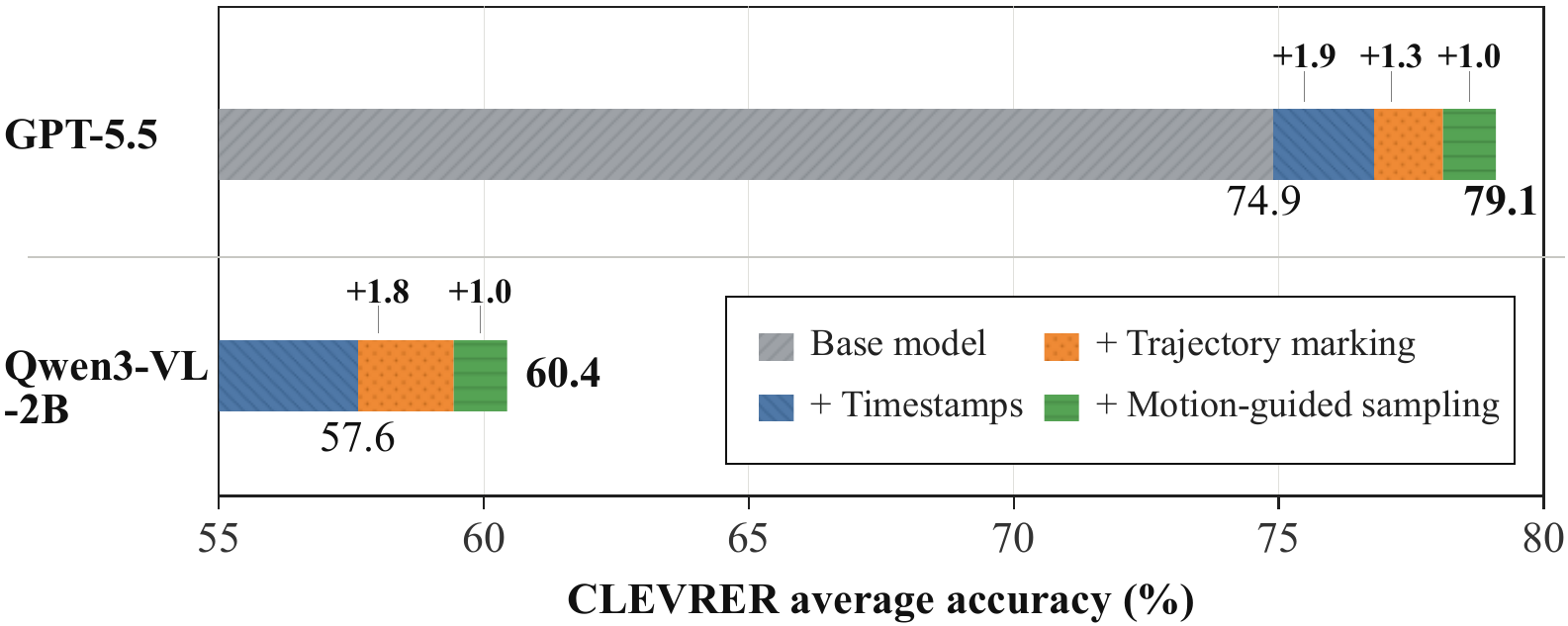}
    \caption{Cumulative ablation of timestamps, trajectory marking, and motion-guided sampling on CLEVRER.}
    \label{fig:ablation}
\end{figure}


To quantify the marginal contribution of each component, we conduct a cumulative ablation on CLEVRER for both models, as shown in Figure~\ref{fig:ablation}. Since timestamp encoding is natively supported by Qwen3-VL but the treatment of timestamps is unclear to GPT-5.5, the two ablation sequences begin from different uniform-sampling baselines. For GPT-5.5, we first add timestamps to the base model, whereas Qwen3-VL-2B starts directly from its native timestamped baseline. We then successively introduce inter-frame trajectory marking and motion-guided sampling.
Trajectory marking improves GPT-5.5 and Qwen3-VL-2B by $+1.3\%$ and $+1.8\%$, respectively, while motion-guided sampling provides additional gains of $+1.0\%$ and $+1.0\%$. The consistent improvements across both models demonstrate that \system's effectiveness comes from explicitly modeling the key motion information. 
More importantly, trajectory marking accounts for the larger share of the total gain, confirming that explicitly exposing inter-frame motion is the primary source of improvement. Motion-guided sampling serves as a complementary component by allocating more of the frame budget to dynamically informative intervals.

\subsection{Preprocessing Cost}

\begin{table}[t]
    \centering
    \small
    \setlength{\tabcolsep}{4pt}
    \begin{tabular}{@{}lrrrcc@{}}
        \toprule
        \multirow{3}{*}{\textbf{Method}}
        & \multicolumn{3}{c}{\textbf{Time (ms)}}
        & \multicolumn{2}{c}{\textbf{Memory (GB)}} \\
        \cmidrule(lr){2-4}\cmidrule(lr){5-6}
        & CLEVRER & SSv2\ \ \ \  & {\scriptsize TempCompass} & \multirow{2}{*}{Avg.} & \multirow{2}{*}{Peak} \\
        & ($\sim$5.12\,s) & ($\sim$4.21\,s) & ($\sim$10.88\,s) & & \\
        \midrule
        AKS & 1022 & 795 & 1924 & 1.70 & 2.70  \\
        FOCUS & 2012 & 1271 & 2678 & 1.69 & 2.70  \\
        SoM & 43203 & 23910 & 47583 & 2.70 & 6.11  \\
        GoM & 2381 & 7057 & 19681 & 5.60 & 9.55 \\
        \textbf{\system} & \textbf{783} & \textbf{609} & \textbf{1920} & \textbf{0.35} & \textbf{3.41} \\
        \bottomrule
    \end{tabular}
    \caption{Per-video preprocessing time and GPU memory overhead of these methods.  ``($\sim$)'' denotes average duration.}
    \label{tab:cost}
\end{table}


We further report the preprocessing latency and GPU memory overhead of each method across the three benchmarks, as shown in Table~\ref{tab:cost}. \system achieves the lowest or comparable preprocessing latency across the three benchmarks, with an average per-video processing time of 783 ms on CLEVRER, 609 ms on SSv2, and 1{,}920 ms on TempCompass.
It is substantially faster than the pixel-level prompting methods SoM and GoM, while remaining competitive with the frame-selection baselines AKS and FOCUS.
\system also maintains low average GPU memory overhead. Owing to its compact 98 MB tracker and window-based streaming tracking, it uses only 0.35 GB on average. Its peak memory is slightly higher than that of selection-only methods because point tracking temporarily allocates large intermediate tensors and incurs allocator-cache fragmentation, but it remains below SoM and GoM. Overall, \system introduces only modest computational, memory, and storage overhead.

\section{Conclusion}

We present Motion-as-Prompt (\system), a training-free, plug-and-play visual-prompting framework that mitigates inter-frame motion loss suffered by MLLMs under sparse sampling in videos. \system recovers inter-frame motion from full-frame-rate video with a frozen point tracker, computes motion energy scores to guide keyframe sampling, and explicitly marks the trajectories between adjacent sampled frames onto the keyframes, so that the MLLM can directly perceive object displacement, direction change, and dynamic interaction. Experiments show that \system consistently improves MLLM performance on motion-reasoning tasks, and causes no degradation on broader video-reasoning tasks.  At a low preprocessing cost, \system enhances the motion understanding of MLLMs, offering a simple and effective visual-prompting augmentation for training-free video motion reasoning.


	
	
	
	
	\bibliography{MaP}
	
\end{document}